\documentclass{article}

\usepackage{graphicx}
\usepackage{subcaption}
\usepackage[final,nonatbib]{tackling_climate_workshop_style}

\usepackage[utf8]{inputenc} 
\usepackage[T1]{fontenc}    
\usepackage[hidelinks]{hyperref}       
\usepackage{url}            
\usepackage{booktabs}       
\usepackage{amsfonts}       
\usepackage{amsmath}
\usepackage{amssymb}
\usepackage{nicefrac}       
\usepackage{microtype}      
\usepackage[capitalize]{cleveref}
\newcommand*{\Scale}[2][4]{\scalebox{#1}{$#2$}}%

\title{Enhanced physics-informed neural networks (PINNs) for high-order power grid dynamics}
\author{%
  Vineet Jagadeesan Nair\thanks{Work done as part of a research internship at X, the moonshot factory (formerly Google X).} \\
  Department of Mechanical Engineering \\
  Massachusetts Institute of Technology\\
  Cambridge, MA 02139 \\
  \texttt{jvineet9@mit.edu} \\
}

\begin{document}

\maketitle

\begin{abstract}
  We develop improved physics-informed neural networks (PINNs) for high-order and high-dimensional power system models described by nonlinear ordinary differential equations. We propose some novel enhancements to improve PINN training and accuracy and also implement several other recently proposed ideas from the literature. We successfully apply these to study the transient dynamics of synchronous generators. We also make progress towards applying PINNs to advanced inverter models. Such enhanced PINNs can allow us to accelerate high-fidelity simulations needed to ensure a stable and reliable renewables-rich future grid.
\end{abstract}

\section{Introduction and motivation}

Rapid decarbonization of power systems is crucial to reducing emissions and enabling clean electrification in other sectors like transportation and industry. This entails a transition away from conventional fossil fuel-based power generation (coal and natural) towards renewables like wind, solar, and batteries. Centralized thermal generators consist of large, rotating synchronous machines with high inertia that help resolve imbalances between power demand and supply to stabilize grid frequency. However, renewable resources and battery storage are largely connected to the grid via inverters based on power electronics with much faster dynamics and have little to no inertia. This introduces reliability and stability challenges for the future grid \cite{dorfler2023control}. Thus, it is essential to be able to accurately study the dynamics of such inverter-based resources (IBRs), along with those of conventional synchronous generators (SGs) which will also continue playing an important role in complementing renewables for the next few decades. This is especially critical during disturbances like faults, outages of lines or generators, and sudden drops in renewables output (e.g. due to cloud cover on solar or wind lulls). Analyzing power system dynamics and transient stability involves solving several nonlinear ordinary differential equations (ODEs), often at very fine resolutions ranging from a few milliseconds to microseconds. These can be expensive to solve using traditional numerical integration, especially for smaller time steps. Furthermore, unlike the relatively small number of centralized power plants today, the increasing penetration of distributed energy resources in the future will significantly increase the number of devices that need to be studied. This motivates the need for more scalable and computationally efficient tools to predict power system dynamics. Scientific machine learning (SciML) based on physics-informed neural networks is a promising data-driven approach to increase the speed of dynamic simulations while still maintaining high accuracy \cite{cuomo2022scientific,raissi2019physics}. This will allow us to accelerate high-fidelity studies of large signal stability of bulk power systems with current-constrained IBRs. Such tools are crucial to accelerate decarbonization while also ensuring a stable and reliable future grid.

\subsection{Prior work and contributions}

A few works in the literature have explored applications of PINNs for power and energy systems \cite{huang2022applications} to directly predict the solutions of nonlinear ODEs. In particular, these have been used to study the dynamics of power systems. Most of these studies have focused on the dynamics of SGs \cite{misyris2020physics,stiasny2023physics}. To our knowledge, only one work so far has considered PINN applications for inverters \cite{nellikkath2024physics}. However, these studies have generally used simplified, relatively low-order, and low-dimensional models. For example, SGs have been studied using the classical, second-order swing equation \cite{kundur2007power} or other reduced-order models \cite{stiasny2024pinnsim}. The inverter studies have focused solely on studying one component, i.e. the phase-locked loop (PLL) converter (which synchronizes the inverter with the grid) by converting the simplified second-order dynamics to an equivalent swing equation with two states. In this work, we aim to build upon existing works and apply PINNs to more complex, higher-order, and higher-dimensional ODE models. In particular, we study a fourth-order SG model with four states and for the inverter, we develop a PINN that models all the components of the inverter (including multiple controllers, converters, and filters) as a whole with the goal of predicting 17 states in total. This increased complexity introduces numerous new challenges for training the PINN and improving its generalization. We provide a comprehensive overview of these difficulties from our experiments as well as observations from prior works. We then attempt to address some of these by proposing some novel enhancements and also by implementing several other ideas from the PINN and broader SciML literature that have not yet been employed in the power systems domain.

\section{Methodology}

\subsection{Physics-informed neural networks (PINNs)}

PINNs are a class of neural networks that aim to directly learn the solutions of differential equations. This is primarily achieved by cleverly designing the loss function during training. Consider the following initial value problem (IVP) with system states $\textbf{x}$, inputs $\textbf{u}$, and initial conditions (ICs) $\textbf{x}_0$. We would like to approximate the solution $\hat{\textbf{x}}(t)$ using a neural network $f_{NN}$ parametrized by $\boldsymbol\theta$, i.e. the weights and biases.
\begin{gather}
    \dot{\textbf{x}} = f(\textbf{x},t,\textbf{u}), \quad \textbf{x}(0) = \textbf{x}_0, \quad \hat{\textbf{x}}(t) = f_{NN}(\textbf{x}_0,t,u\;\boldsymbol\theta) \label{eq:ode}
\end{gather}
We train the PINN by minimizing a weighted combination of two types of losses: (i) $\mathcal{L}_{ic}$ that enforces the ICs, and (ii) the ODE residual $\mathcal{L}_{ode}$ that enforces the ODE dynamics throughout the time domain:
\begin{gather}
    \min_{\boldsymbol\theta} \mathcal{L}(\boldsymbol{\theta}) = \lambda_{ic} \mathcal{L}_{ic} + \lambda_{ode} \mathcal{L}_{ode}, \quad \mathcal{L}_{ic} = \frac{\|\textbf{x}_0 - \hat{\textbf{x}}(0)\|}{n_{ic}}, \quad \mathcal{L}_{ode} = \frac{\|\dot{\hat{\textbf{x}}}(t) - f(\hat{\textbf{x}},t,\textbf{u})\|}{n_{ode}} \label{eq:pinn}
\end{gather}
where $n_{ic}$ is the number of training ICs and $n_{ode}$ is the number of collocation points chosen throughout the simulation time period $[0,t_{end}]$ at which we enforce the ODE. We found that randomly sampling these points provided better results than using equally spaced points in the time domain \cite{wang2023expert}. We used the averaged $L_2$ norm for both losses. Note that prior work has shown the $L_{\infty}$ norm may also be a suitable choice \cite{wang20222} in some cases. With just these two terms, we can train our PINN in an unsupervised fashion that does not require us to solve the ODE beforehand with numerical integration. In a supervised setting, we can also include a third term that penalizes deviations of the PINN solution from the ground truth $\textbf{x}^*(t)$, i.e. $\mathcal{L}_{sol} = \frac{\|\hat{\textbf{x}}(t) - \textbf{x}^*(t)\|}{n_{sol}}$. However, this requires solving the ODE offline to generate ground truth trajectories. The overall loss function can then be minimized using gradient descent and backpropagation. The gradients of the predicted PINN solutions (for $\mathcal{L}_{ode}$) can be efficiently computed via automatic differentiation. Both the SG and inverter models are nonlinear state space models of the form: $\dot{\textbf{x}}(t) = f(\textbf{x},t,\textbf{u}) = A \textbf{x}(t) + R(\textbf{x},\textbf{u})$, where the matrix $R$ captures the nonlinearities in the model. Further details on these models can be found in \cref{app:models}.

\subsection{Overview of challenges with PINNs}

PINNs are difficult to train, especially for higher-order, stiff ODE systems. One major issue is the ill-conditioning of the loss function, particularly near the optimal solution \cite{krishnapriyan2021characterizing,rathore2024challenges}. We found this to be true for our experiments too, by inspecting the condition number of the loss Hessian. Previous works have shown that this is due to the differential operator in the ODE residual term, resulting in many local minima and saddle points in the loss landscape that are difficult to escape, leading to poor convergence and sometimes instability. This issue is exacerbated in the case of higher-order models like those considered here, and the curse of dimensionality also makes training and inference computationally expensive for ODEs and PDEs with three or more states \cite{hu2024tackling}. Another challenge is balancing the different goals during training, namely satisfying initial conditions while also learning the physical dynamics described by the ODE. These significantly affect PINN generalization.  

\subsection{Proposed enhancements to improve the PINN performance}

\paragraph{Tuning neural network architecture and hyperparameters:}
We used grid search hyperparameter tuning to determine a good neural network size. For \texttt{sgPINN}, we used 4 dense, multilayer perception (MLP) layers and 64 neurons per layer. For \texttt{invPINN}, we used a slightly larger network with 5 layers with 128 neurons each, due to the higher model complexity. Both used $tanh$ activation functions, Glorot (or Xavier) uniform initialization for all weights \cite{glorot2010understanding} and biases initialized as zero. We sampled 50 initial conditions within the stable regime and 1000 collocation points. We found that combining both ADAM and L-BFGS optimizers obtained the best results, allowing us to drive the loss down further. We first ran ADAM (for 20,000 iterations) with an initial learning rate of 0.01 and an exponential decay factor of 0.9 applied every 100 epochs. This was followed by L-BFGS for another 5,000 iterations with the default step size of 1.0 and strong Wolfe line search. This result is also in line with other works \cite{zubov2021neuralpde} that showed that utilizing quasi-Newton methods like L-BGFS can be helpful after the loss value stagnates with gradient-based methods like ADAM. Although first-order methods can avoid saddle points, they converge slowly when the function is ill-conditioned. However, second-order methods like L-BFGS also leverage Hessian information and approximate the loss function as quadratic when close to the optimum, thus locally improving the conditioning \cite{rathore2024challenges}. As part of future work, we also aim to extend this approach by experimenting with third-order Chebyshev-Halley methods to achieve even faster convergence \cite{gundersen2010large}.

\paragraph{Multiobjective optimization:}
One major challenge with training PINNs is balancing the different terms in the linear weighted sum loss function from \cref{eq:pinn}, which may be of very different magnitudes. In our case, we found that the ODE residual generally dominates the loss function over the IC loss. Thus, we would ideally like to prioritize driving down the ODE loss first and then switch to focus more on the IC loss in later training stages. Correctly setting the relative weights ($\lambda_{ic},\lambda_{ode}$) on these loss terms can heavily influence the training process and help bias toward learning solutions that satisfy both ICs and the ODE. Inspired by a popular technique in multiobjective optimization \cite{grodzevich2006normalization}, we propose normalizing the objective function by the differences in the optimal values at the \textit{Nadir} and \textit{Utopia} points, which gives us the length of the intervals where the optimal loss values vary within the Pareto optimal set. The \textit{Utopia} point is the ideal objective vector $[\mathcal{L}_{ic}^U, \mathcal{L}_{ode}^U]$ obtained by minimizing each individual loss function (i.e. either the IC loss or the ODE loss) separately, i.e. $\mathcal{L}^U_i = \mathcal{L}_i(\boldsymbol{\theta}_i^U) = \min_{\boldsymbol\theta} \mathcal{L}_i(\boldsymbol\theta), \; i \in \{ic,ode\}$. This point normally isn't feasible because the two objectives of satisfying the ICs and the ODE are often conflicting, but it does provide lower bounds for the Pareto set. The \textit{Nadir} point on the other hand, provides an upper bound, with the objective vector given by $[\mathcal{L}_{ic}^N, \mathcal{L}_{ode}^N]$, where $\mathcal{L}^N_i = \max_{j\in \{ic,ode\}} \mathcal{L}_i(\boldsymbol{\theta}_j^U)$. We can then normalize the two loss terms by initializing the regularization weights as $\lambda_i = \frac{1}{\mathcal{L}^N_i - \mathcal{L}^U_i} \; \forall i \in \{ic,ode\}$, thus ensuring that both losses are similar in magnitude. In our case, we can approximate these \textit{Utopia} and \textit{Nadir} points by first training two smaller auxiliary networks that optimize the IC and ODE losses, respectively. We train both for a sufficient number of epochs until the loss plateaus at a low value.

\paragraph{Adaptive balancing of different loss gradients:}
After initializing the loss weights as above, we can further also adaptively tune these weights during training based on the intermediate values of the gradients of the individual loss terms \cite{wang2023expert}. This ensures that both the weighted IC and ODE loss gradients have similar norms to avoid biasing excessively to minimize one term at the expense of the other. Prior works have also proposed other approaches to achieve similar balancing, based on Gaussian processes \cite{mcclenny2023self,xiang2022self} and neural tangent kernels \cite{wang2022and}.

\paragraph{Sequence-to-sequence learning:}
Instead of trying to predict ODE solutions for the whole time domain at once, we also experimented with the approach of training the PINN to only predict the very next time step, which is likely an easier task \cite{krishnapriyan2021characterizing}. We can then use the PINN predictions for the next timestep $\hat{\textbf{x}}(\Delta t)$ as an IC for predicting the solution for the following timestep $\hat{\textbf{x}}(2\Delta t)$. In our case, we found that this provided some marginal benefits in improving the \texttt{sgPINN} performance. We expect this to be especially useful in further reducing errors for \texttt{invPINN} due to its faster dynamics. 

Although not required for the proposal, we have included in \cref{app:sim} preliminary simulation results that show the good performance of \texttt{sgPINN} and shed some insights into the unique challenges posed by \texttt{invPINN}. Furthermore, in \cref{app:future}, we discuss the limitations of our approach (based on observations from our experiments so far), some ongoing efforts to address these challenges as well as several areas for future work.


\section{System models\label{app:models}}
\subsection{Synchronous generator model}
For the SG, we used a two-axis model for a generator connected to an described in \cite{natarajan2014method,wang2015power}, which is a nonlinear fourth-order ODE as in \cref{eq:sg_model} with parameters shown in \cref{tab:sg_params}. Using the \texttt{sgPINN}, we aim to predict all four states. These represent the generator currents in the $d\text{-}q$ coordinates ($i_d$, $i_q$), the generator frequency $\omega = \dot{\theta}$ and angle difference $\delta = \theta - \theta_g - \frac{\pi}{2}$, where $\theta_g, \theta$ are the absolute grid voltage angle and rotor angle, respectively.

\begin{gather}
    \left[\begin{array}{c} \dot{i_d} \\ \dot{i_q} \\ \dot{\omega} \\ \dot{\delta}\end{array}\right]=\left[\begin{array}{cccc}-\frac{R_s}{L_s} & \omega & 0 & 0 \\ -\omega & -\frac{R_s}{L_s} & -\frac{m i_f}{L_s} & 0 \\ 0 & \frac{m i_f}{J} & -\frac{D_p}{J} & 0 \\ 0 & 0 & 1 & 0\end{array}\right]\left[\begin{array}{c}i_d \\ i_q \\ \omega \\ \delta\end{array}\right]+\left[\begin{array}{c}\frac{V}{L_s} \sin \delta \\ \frac{V}{L_s} \cos \delta \\ \frac{T_m}{J} \\ -\omega_g\end{array}\right] \label{eq:sg_model}
\end{gather}

\begin{table}[htb]
\centering
\begin{tabular}{llc}
\toprule
\textbf{Parameter} & \textbf{Symbol} & \textbf{Value} \\
\midrule
Nominal Grid Frequency & $\omega_g$ & $100\pi$ rad/sec \\
Stator Resistance & $R_s$ & 0.152 $\Omega$ \\
Stator Inductance & $L_s$ & 4.4 mH \\
Damping Coefficient & $D_p$ & 10.14 Nm/(rad/sec) \\
Inertia Constant & $J$ & 0.02 Kgm$^2$/rad \\
Mechanical Torque & $T_m$ & [15.9 + $D_p\omega_g$] Nm \\
Nominal Voltage & $V$ & $230\sqrt{3}$ Volts \\
Field Excitation Constant & $m_{if}$ & $-1.38$ Voltsec \\
\bottomrule
\end{tabular}
\caption{Nominal parameters for a 5kW SG.\label{tab:sg_params}}
\end{table}

\subsection{Inverter model}

We used a nonlinear, state-space inverter model for grid-connected inverters under current control mode, developed in \cite{kroutikova2007state}. This describes grid-following (GFL) inverters that act as a current source to inject active and reactive power into the grid. They simply synchronize with the grid but cannot support grid voltage or frequency. Although GFL inverters are most common today, grid-forming inverters (GFM) will become more important in the future with the increasing presence of IBRs. These act as voltage sources and can thus help support the grid voltage and frequency. We will also extend our \texttt{invPINN} to consider GFM models as part of future work \cite{du2020modeling}. The GFM inverter model consists of 17 states. The output voltages ($v_{Oabc}$ or $v_{Odq}$) are determined by the nominal phase voltage of the grid. 
\[\Scale[0.9]{
    \mathbf{x} = \left[ \theta \quad \Phi_{\text{PLL}} \quad i_{Ld}^{*} \quad i_{Lq}^{*} \quad q_{3Ld} \quad q_{3Lq} \quad q_{Ld}^{\text{err}} \quad q_{Lq}^{\text{err}} \quad i_{Ld} \quad i_{Lq} \quad i_{LO} \quad v_{Cd} \quad v_{Cq} \quad v_{CO} \quad i_{Od} \quad i_{Oq} \quad i_{OO} \right]^{T}
}\]

The matrices describing the model $\dot{\textbf{x}}(t) = A \textbf{x}(t) + R(\textbf{x},\textbf{u})$ are given by:
\[\Scale[0.8]{A = \left[
\begin{array}{ccccccccccccccccccc}
0 & K_I^{PLL} & 0 & 0 & 0 & 0 & 0 & 0 & 0 & 0 & 0 & 0 & 0 & 0 & 0 & 0 & 0 & 0 & 0 \\
0 & 0 & 0 & 0 & 0 & 0 & -\sqrt{2}\omega_c & 0 & 0 & 0 & 0 & 0 & 0 & 0 & 0 & 0 & 0 & 0 & 0 \\
0 & 0 & 0 & 0 & 0 & 0 & 0 & \omega_c^2 & 0 & 0 & 0 & 0 & 0 & 0 & 0 & 0 & 0 & 0 & 0 \\
0 & 0 & 0 & 0 & 0 & 0 & -\sqrt{2}\omega_c & 0 & 0 & 0 & 0 & 0 & 0 & 0 & 0 & 0 & 0 & 0 & 0 \\
0 & 0 & 0 & 0 & 0 & 0 & 0 & \omega_c^2 & 0 & 0 & 0 & 0 & 0 & 0 & 0 & 0 & 0 & 0 & 0 \\
0 & 0 & 0 & -1 & 0 & 0 & 0 & 0 & 0 & 0 & 0 & 0 & 0 & 0 & 0 & -1 & 0 & 0 & 0 \\
0 & 0 & 0 & 0 & -1 & 0 & 0 & 0 & 0 & 0 & 0 & 0 & 0 & 0 & 0 & 0 & -1 & 0 & 0 \\
0 & 0 & 1 & 0 & 0 & -1 & 0 & 0 & 0 & 0 & 0 & 0 & 0 & 0 & 0 & 0 & 0 & 0 & 0 \\
0 & 0 & 0 & 0 & 1 & 0 & 0 & 0 & 0 & 0 & 0 & 0 & 0 & 0 & 0 & 0 & 0 & 0 & 0 \\
0 & 0 & 1 & 0 & 0 & 0 & 0 & 0 & 0 & 0 & 0 & 0 & 0 & 0 & 0 & 0 & 0 & 0 & 0 \\
0 & 0 & 0 & 0 & 0 & -1 & 0 & 0 & 0 & 0 & 0 & 0 & 0 & 0 & 0 & 0 & 0 & 0 & 0 \\
0 & 0 & 0 & 0 & 0 & 0 & 0 & 0 & -\frac{R}{L} & 0 & 0 & 0 & 0 & 0 & 0 & 0 & 0 & 0 & 0 \\
0 & 0 & 0 & 0 & 0 & 0 & 0 & 0 & 0 & -\frac{R}{L} & 0 & 0 & 0 & 0 & 0 & 0 & 0 & 0 & 0 \\
0 & 0 & 0 & 0 & 0 & 0 & 0 & 0 & 0 & 0 & -\frac{1}{L} & 0 & 0 & 0 & 0 & 0 & 0 & 0 & 0 \\
0 & 0 & 0 & 0 & 0 & 0 & 0 & 0 & 0 & 0 & 0 & -\frac{1}{L} & 0 & 0 & 0 & 0 & 0 & 0 & 0 \\
0 & 0 & 0 & 0 & 0 & 0 & 0 & 0 & -\frac{1}{C} & 0 & 0 & 0 & 0 & 0 & 0 & 0 & 0 & 0 & 0 \\
0 & 0 & 0 & 0 & 0 & 0 & 0 & 0 & 0 & -\frac{1}{C} & 0 & 0 & 0 & 0 & 0 & 0 & 0 & 0 & 0 \\
0 & 0 & 0 & 0 & 0 & 0 & 0 & 0 & \frac{1}{C} & 0 & 0 & 0 & 0 & 0 & 0 & 0 & 0 & 0 & 0 \\
0 & 0 & 0 & 0 & 0 & 0 & 0 & 0 & 0 & \frac{1}{C} & 0 & 0 & 0 & 0 & 0 & 0 & 0 & 0 & 0 \\
0 & 0 & 0 & 0 & 0 & 0 & 0 & 0 & 0 & 0 & -\frac{R_{\text{coupl}}}{L_{\text{coupl}}} & 0 & 0 & 0 & 0 & 0 & 0 & 0 & 0 \\
0 & 0 & 0 & 0 & 0 & 0 & 0 & 0 & 0 & 0 & 0 & -\frac{R_{\text{coupl}}}{L_{\text{coupl}}} & 0 & 0 & 0 & 0 & 0 & 0 & 0 \\
0 & 0 & 0 & 0 & 0 & 0 & 0 & 0 & -\frac{1}{C} & 0 & 0 & 0 & 0 & 0 & 0 & 0 & 0 & 0 & 0 \\
0 & 0 & 0 & 0 & 0 & 0 & 0 & 0 & 0 & -\frac{1}{C} & 0 & 0 & 0 & 0 & 0 & 0 & 0 & 0 & 0 \\
0 & 0 & 0 & 0 & 0 & 0 & 0 & 0 & \frac{1}{C} & 0 & 0 & 0 & 0 & 0 & 0 & 0 & 0 & 0 & 0 \\
0 & 0 & 0 & 0 & 0 & 0 & 0 & 0 & 0 & \frac{1}{C} & 0 & 0 & 0 & 0 & 0 & 0 & 0 & 0 & 0 \\
0 & 0 & 0 & 0 & 0 & 0 & 0 & 0 & 0 & 0 & -\frac{R_{\text{coupl}}}{L_{\text{coupl}}} & 0 & 0 & 0 & 0 & 0 & 0 & 0 & 0 \\
0 & 0 & 0 & 0 & 0 & 0 & 0 & 0 & 0 & 0 & 0 & -\frac{R_{\text{coupl}}}{L_{\text{coupl}}} & 0 & 0 & 0 & 0 & 0 & 0 & 0 \\
0 & 0 & 0 & 0 & 0 & 0 & 0 & 0 & -\frac{3R_G - R_{\text{coupl}}}{L_{\text{coupl}}} & 0 & 0 & 0 & 0 & 0 & 0 & 0 & 0 & 0 & 0 \\
0 & 0 & 0 & 0 & 0 & 0 & 0 & 0 & 0 & -\frac{3R_G - R_{\text{coupl}}}{L_{\text{coupl}}} & 0 & 0 & 0 & 0 & 0 & 0 & 0 & 0 & 0 \\
\end{array} \right]}\]

\[\Scale[0.8]{R(\textbf{x}, \textbf{u}) = \left[
\begin{array}{c}
K_P^{\text{PLL}} v_{Oq} \\
v_{Oq} \\
0 \\
0 \\
\frac{v_{Od}P^* - v_{Oq}Q^*}{v_{Od}^2 + v_{Oq}^2} \\
\frac{v_{Oq}P^* + v_{Od}Q^*}{v_{Od}^2 + v_{Oq}^2} \\
0 \\
0 \\
\frac{1}{L} v_{1d} + \omega i_{Lq} \\
\frac{1}{L} v_{1q} - \omega i_{Ld} \\
\omega v_{Cq} \\
-\omega v_{Cd} \\
-\frac{1}{L_{\text{coupl}}} v_{Od} + \omega i_{Oq} \\
-\frac{1}{L_{\text{coupl}}} v_{Oq} - \omega i_{Od} \\
\end{array}
\right]}\]

\begin{figure}[htb]
  \centering
  \begin{subfigure}[b]{0.49\linewidth}
  \includegraphics[width=\linewidth]{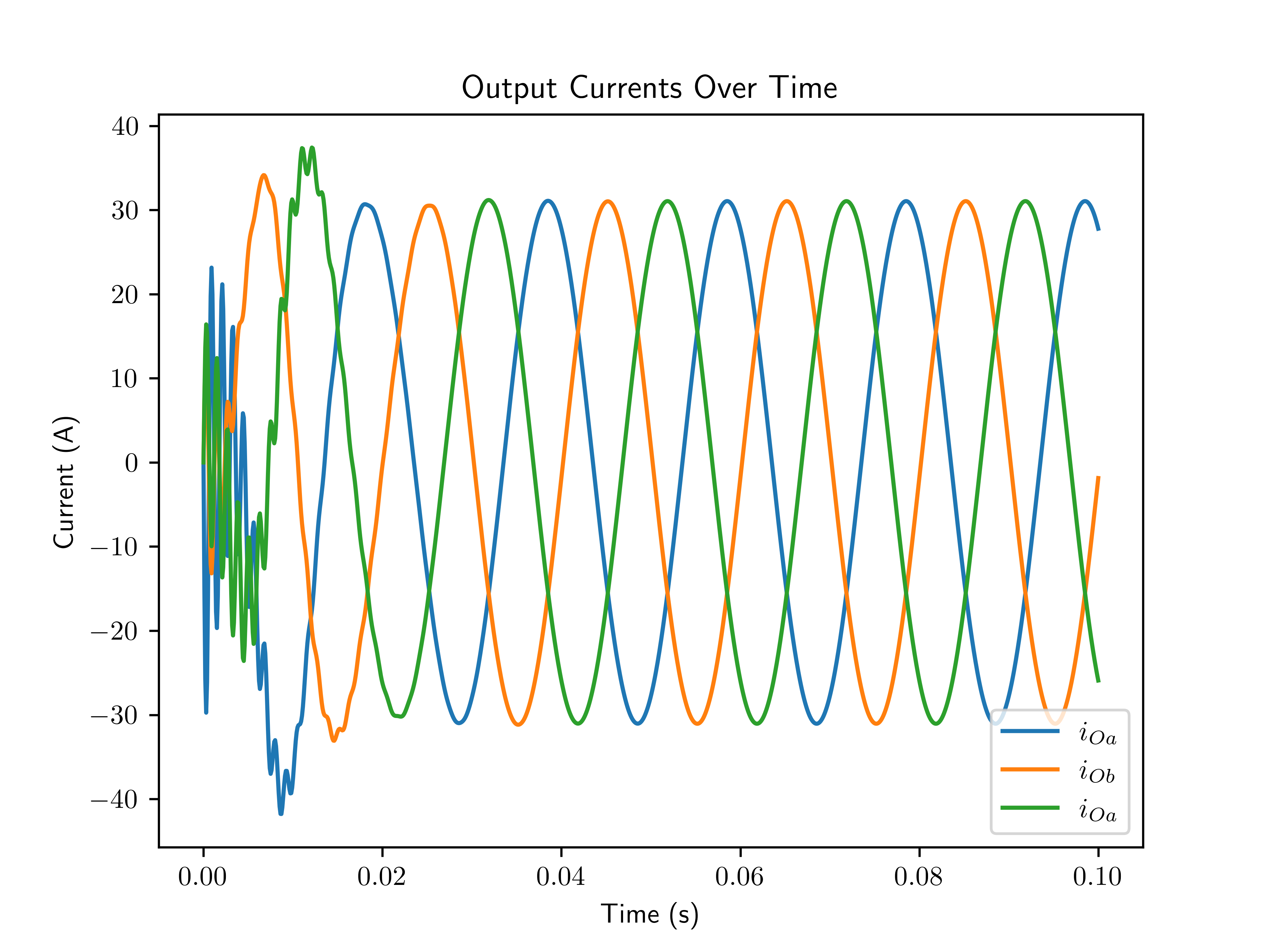}
  \caption{Inverter outputs on 3 phases A, B, C.}
  \end{subfigure}
  \begin{subfigure}[b]{0.49\linewidth}
  \includegraphics[width=\linewidth]{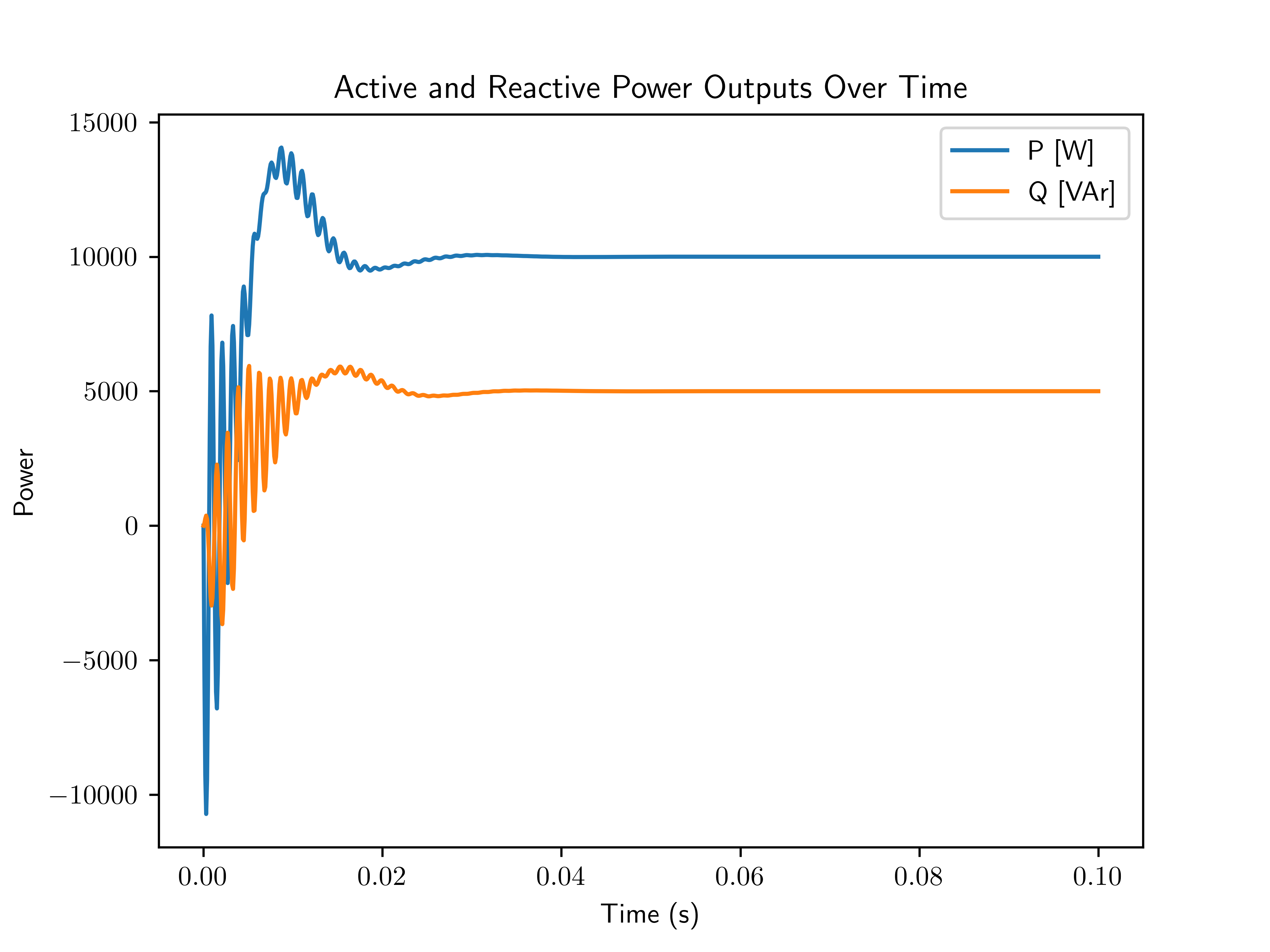}
  \caption{Inverter output active and reactive power.}
  \end{subfigure}
  \caption{Inverter simulation results highlighting faster dynamics that are more challenging to capture, than with SGs.\label{fig:inv_sims}}
\end{figure}

This model includes all the main components of the GFL converter, namely the current power controller, current controller, phase-locked loop, LC filter, and coupling impedance, along with the transformations from the $d\text{-}q$ to $a\text{-}b\text{-}c$ coordinates. Note that in addition to the state space formulation shown above, the model also implicitly includes another important nonlinearity while calculating $v_{Id}$ and $v_{Iq}$ with a possible saturation limit. For more details, please refer to \cite{kroutikova2007state}. \cref{tab:inv_params} lists the main parameters used for our simulations. Note that we had to tune and modify some values (compared to the original paper) to obtain stable simulation results. The main outputs of interest are the output currents and power injections from the inverter. The 3-phase ($a\text{-}b\text{-}c$) currents shown in \cref{fig:inv_sims} can be readily derived from the $d\text{-}q$ currents which are part of the inverter's state.

\begin{table}[htb]
\centering
\begin{tabular}{ll}
\toprule
\textbf{Parameter} & \textbf{Value} \\
\midrule
Nominal phase voltage & 240 V \\
Grid frequency & 50 Hz \\
Coupling impedance $R_{coupl}, \; L_{coupl}$ & $(0.131 + j0.96) \ \Omega$ \\
DC bus voltage & 1000 V \\
\midrule
\textbf{Coupling filter} & \\
\quad Inductance $L$ & 1.35 mH \\
\quad Capacitance $C$ & 50 $\mu$F \\
\quad Resistance $R$ & 0.056 $\Omega$ \\
\quad Ground resistance $R_G$ & 100 $\Omega$ \\
\midrule
Switching frequency $\omega_c$ & 100 Hz \\
Reference active power $P^*$ & 10 kW \\
Reference reactive power $Q^*$ & 5 kVar \\
\midrule
\textbf{Current controller} & \\
\quad Proportional gain in branch $d$ $K_P^d$ & 1 \\
\quad Integral gain in branch $d$ $K_I^d$ & 460 \\
\quad Proportional gain in branch $q$ $K_P^q$ & 1 \\
\quad Integral gain in branch $q$ $K_I^q$ & 460 \\
\midrule
\textbf{PLL} & \\
\quad Proportional gain $K_P^{PLL}$ & 2.1 \\
\quad Integral gain $K_I^{PLL}$ & 5000 \\
\bottomrule
\end{tabular}
\caption{Parameter values used for inverter simulations and \texttt{invPINN}.\label{tab:inv_params}}
\end{table}

\section{Preliminary simulation results \label{app:sim}}

\begin{figure}[htb]
  \centering
  \begin{subfigure}[b]{0.49\linewidth}
  \includegraphics[width=\linewidth]{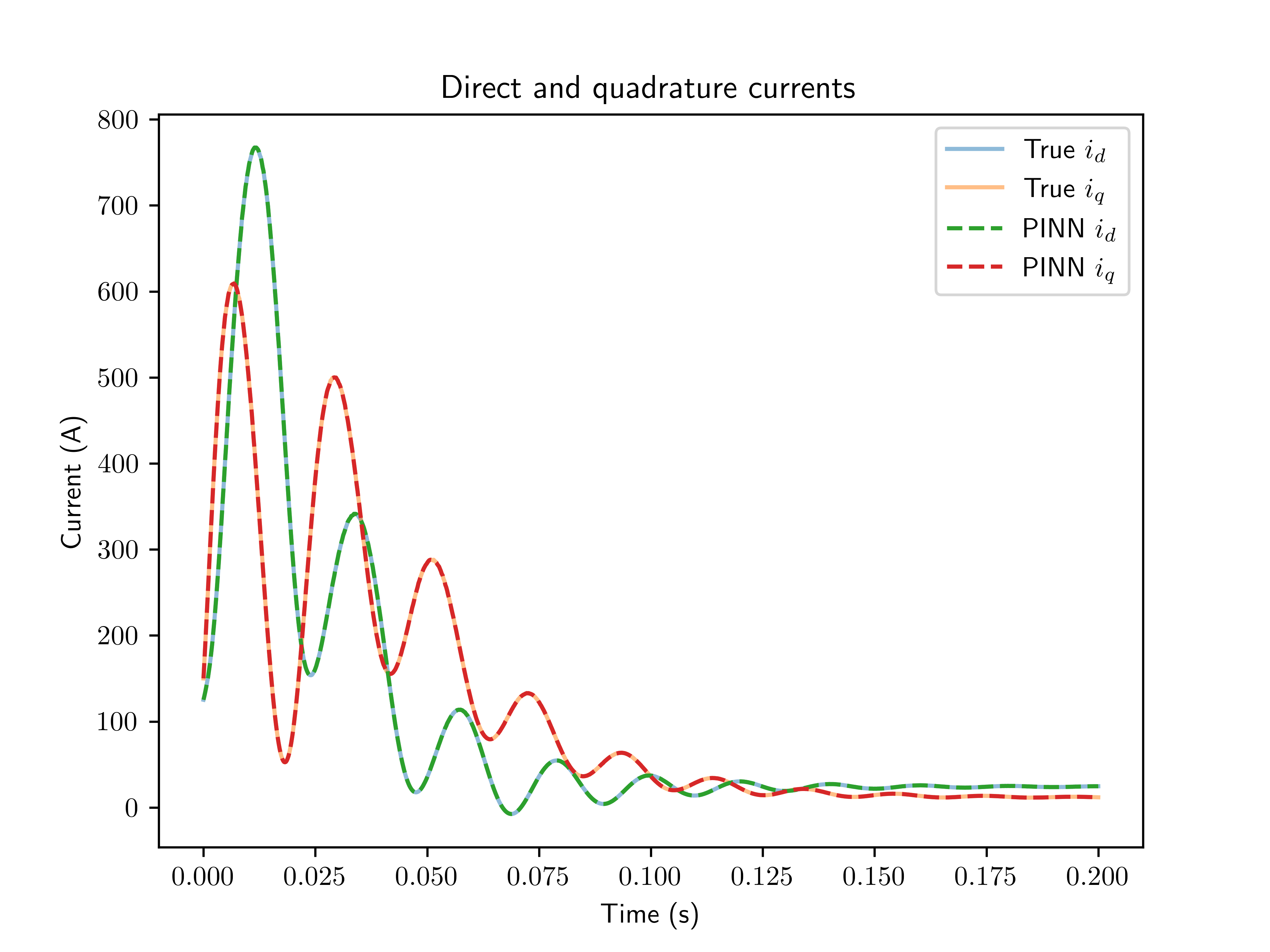}
  \caption{Direct-axis and quadrature-axis currents.}
  \end{subfigure}
  \begin{subfigure}[b]{0.49\linewidth}
  \includegraphics[width=\linewidth]{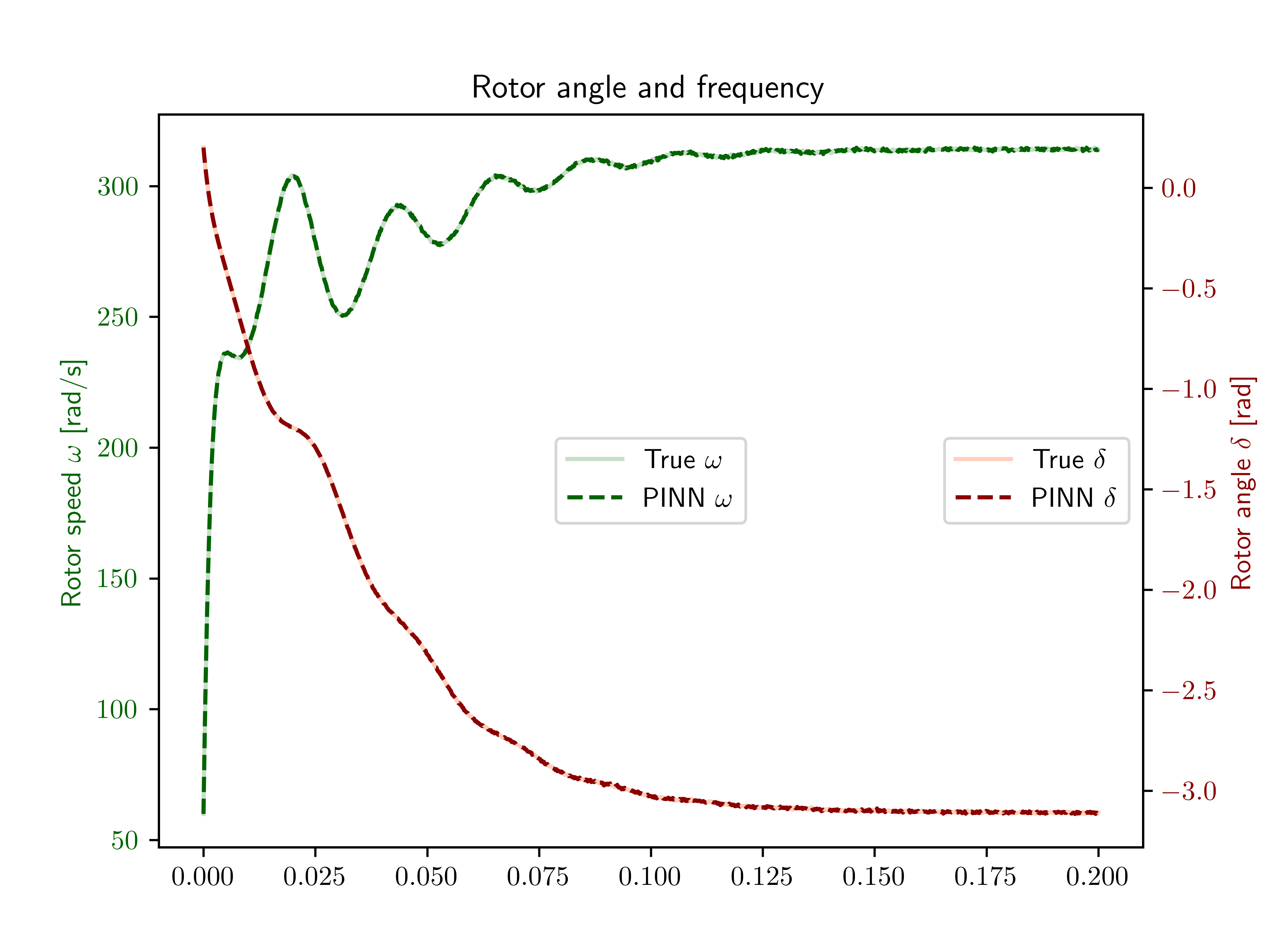}
  \caption{Rotor angle and frequency.}
  \end{subfigure}
  \caption{Comparison of PINN predictions and true solutions for one test initial condition.\label{fig:sg_pinn}}
\end{figure}

We trained both PINNs on Intel Xeon Gold 628 processors with NVIDIA Volta V100 GPUs on the MIT Supercloud high-performance computing cluster \cite{reuther2018interactive}. Here, we provide some preliminary results for \texttt{sgPINN} in \cref{fig:sg_pinn}. We see that \texttt{sgPINN} is able to predict all four states fairly accurately, including the generator currents as well as the rotor angle and angular speed. However, training the \texttt{invPINN} proved to be more challenging. Even after implementing several of the proposed enhancements, the PINN still has quite high prediction errors and does not generalize well. This may be due to the higher order and dimensionality of the state space, and greater nonlinearities - we are currently working on improving \texttt{invPINN} performance (see \cref{app:future}). We have not included the unsatisfactory \texttt{invPINN} predictions here but some sample simulations for the inverter model using numerical integration are shown in \cref{app:inv_sims}. These show how the inverter dynamics (with smaller electrical time constants) are much faster than those of the SG (with larger mechanical time constants). While the SG transients in \cref{fig:sg_pinn} take around 100-200 milliseconds to settle down, the inverter transients in \cref{fig:inv_sims} settle down in less than 10-20 milliseconds, after which the output powers track their reference setpoints. Under certain conditions and disturbances, such inverters can have even faster transients on the order of microseconds. In addition to fast timescales, the inverter ODEs are stiffer and require careful parameter tuning to avoid numerical instability. 

\subsection{Inverter model simulation results \label{app:inv_sims}}

\section{Limitations and future work \label{app:future}}

There are several areas that we are actively exploring to extend this work. \texttt{sgPINN} is able to predict the dynamics reasonably well for a broad range of initial conditions while fixing the inputs and model parameters. However, we face challenges with generalization and prediction accuracy if the test points are far outside the training dataset regime. It is also challenging to sample more diverse initial conditions since certain combinations of states and inputs can potentially lead to instabilities. Extrapolating to predict the dynamics for outlier cases will likely also require more complex and larger NN architectures. We are also currently experimenting with several approaches to improve the performance of \texttt{invPINN}. Firstly, we would like to explore training our PINNs in a supervised manner by also feeding in ground truth solution trajectories during training. Although this could increase the risk of overfitting, utilizing such time-series inputs would also allow us to effectively leverage other types of layers (beyond fully connected MLPs) that are well-suited to model temporal dependencies, such as recurrent, long short-term memory, and even potentially transformer networks. It would also be interesting to consider use cases for generative and large language models in the context of PINNs and power system dynamics \cite{majumder2024exploring,chen2024gpt}. 

We are also interested in applying enhanced PINNs to estimate unknown or uncertain ODE parameters to capture unmodeled or higher-order dynamics and further improve accuracy \cite{raissi2017physics,misyris2021capturing,stiasny2021physics}. This could be done by leveraging realistic data from phasor-measurement units (PMUs) \cite{naglic2019pmu}, as opposed to relying solely on synthetic data generation. However, this also introduces challenges while dealing with noisy, incomplete datasets. We have only validated our \texttt{sgPINN} and \texttt{invPINN} for standalone, single-machine simulations thus far. However, with some fine-tuning and retraining, we can also employ these for larger-scale simulations on realistic networks while considering grid physics and power flow constraints as well as interactions among multiple SGs and IBRs together. Finally, further work is needed to improve the interpretability of black-box PINNs so as to encourage their utilization for safety-critical systems like the power grid \cite{karniadakis2021physics}. This includes improving uncertainty quantification with Bayesian PINNs \cite{yang2021bv2,stock2024bayesian}. Such efforts could also open doors to utilizing PINNs for other important differential equations involved in modeling power and systems \cite{radisavljevic2023modeling}.

\section{Acknowledgements}

The author thanks Hamed Khalilinia and Amir Saadat for valuable guidance and feedback during the project. The author also acknowledges the MIT SuperCloud and Lincoln Laboratory Supercomputing Center for providing (HPC, database, consultation) resources that have contributed to the research results reported within this paper. 

\bibliography{refs_manual}

\begin{thebibliography}{10}

\bibitem{dorfler2023control}
F.~D{\"o}rfler and D.~Gro{\ss}, ``Control of low-inertia power systems,'' {\em Annual Review of Control, Robotics, and Autonomous Systems}, vol.~6, no.~1, pp.~415--445, 2023.

\bibitem{cuomo2022scientific}
S.~Cuomo, V.~S. Di~Cola, F.~Giampaolo, G.~Rozza, M.~Raissi, and F.~Piccialli, ``Scientific machine learning through physics--informed neural networks: Where we are and what’s next,'' {\em Journal of Scientific Computing}, vol.~92, no.~3, p.~88, 2022.

\bibitem{raissi2019physics}
M.~Raissi, P.~Perdikaris, and G.~E. Karniadakis, ``Physics-informed neural networks: A deep learning framework for solving forward and inverse problems involving nonlinear partial differential equations,'' {\em Journal of Computational physics}, vol.~378, pp.~686--707, 2019.

\bibitem{huang2022applications}
B.~Huang and J.~Wang, ``Applications of physics-informed neural networks in power systems-a review,'' {\em IEEE Transactions on Power Systems}, vol.~38, no.~1, pp.~572--588, 2022.

\bibitem{misyris2020physics}
G.~S. Misyris, A.~Venzke, and S.~Chatzivasileiadis, ``Physics-informed neural networks for power systems,'' in {\em 2020 IEEE power \& energy society general meeting (PESGM)}, pp.~1--5, IEEE, 2020.

\bibitem{stiasny2023physics}
J.~Stiasny and S.~Chatzivasileiadis, ``Physics-informed neural networks for time-domain simulations: Accuracy, computational cost, and flexibility,'' {\em Electric Power Systems Research}, vol.~224, p.~109748, 2023.

\bibitem{nellikkath2024physics}
R.~Nellikkath, I.~Murzakhanov, S.~Chatzivasileiadis, A.~Venzke, and M.~K. Bakhshizadeh, ``Physics-informed neural networks for phase locked loop transient stability assessment,'' {\em Electric Power Systems Research}, vol.~236, p.~110790, 2024.

\bibitem{kundur2007power}
P.~Kundur, ``Power system stability,'' {\em Power system stability and control}, vol.~10, pp.~7--1, 2007.

\bibitem{stiasny2024pinnsim}
J.~Stiasny, B.~Zhang, and S.~Chatzivasileiadis, ``Pinnsim: A simulator for power system dynamics based on physics-informed neural networks,'' {\em Electric Power Systems Research}, vol.~235, p.~110796, 2024.

\bibitem{wang2023expert}
S.~Wang, S.~Sankaran, H.~Wang, and P.~Perdikaris, ``An expert's guide to training physics-informed neural networks,'' {\em arXiv preprint arXiv:2308.08468}, 2023.

\bibitem{wang20222}
C.~Wang, S.~Li, D.~He, and L.~Wang, ``Is $l^{2}$ physics informed loss always suitable for training physics informed neural network?,'' {\em Advances in Neural Information Processing Systems}, vol.~35, pp.~8278--8290, 2022.

\bibitem{krishnapriyan2021characterizing}
A.~Krishnapriyan, A.~Gholami, S.~Zhe, R.~Kirby, and M.~W. Mahoney, ``Characterizing possible failure modes in physics-informed neural networks,'' {\em Advances in neural information processing systems}, vol.~34, pp.~26548--26560, 2021.

\bibitem{rathore2024challenges}
P.~Rathore, W.~Lei, Z.~Frangella, L.~Lu, and M.~Udell, ``Challenges in training pinns: A loss landscape perspective,'' {\em arXiv preprint arXiv:2402.01868}, 2024.

\bibitem{hu2024tackling}
Z.~Hu, K.~Shukla, G.~E. Karniadakis, and K.~Kawaguchi, ``Tackling the curse of dimensionality with physics-informed neural networks,'' {\em Neural Networks}, vol.~176, p.~106369, 2024.

\bibitem{glorot2010understanding}
X.~Glorot and Y.~Bengio, ``Understanding the difficulty of training deep feedforward neural networks,'' in {\em Proceedings of the thirteenth international conference on artificial intelligence and statistics}, pp.~249--256, JMLR Workshop and Conference Proceedings, 2010.

\bibitem{zubov2021neuralpde}
K.~Zubov, Z.~McCarthy, Y.~Ma, F.~Calisto, V.~Pagliarino, S.~Azeglio, L.~Bottero, E.~Luj{\'a}n, V.~Sulzer, A.~Bharambe, {\em et~al.}, ``Neuralpde: Automating physics-informed neural networks (pinns) with error approximations,'' {\em arXiv preprint arXiv:2107.09443}, 2021.

\bibitem{gundersen2010large}
G.~Gundersen and T.~Steihaug, ``On large-scale unconstrained optimization problems and higher order methods,'' {\em Optimization Methods \& Software}, vol.~25, no.~3, pp.~337--358, 2010.

\bibitem{grodzevich2006normalization}
O.~Grodzevich and O.~Romanko, ``Normalization and other topics in multi-objective optimization,'' 2006.

\bibitem{mcclenny2023self}
L.~D. McClenny and U.~M. Braga-Neto, ``Self-adaptive physics-informed neural networks,'' {\em Journal of Computational Physics}, vol.~474, p.~111722, 2023.

\bibitem{xiang2022self}
Z.~Xiang, W.~Peng, X.~Liu, and W.~Yao, ``Self-adaptive loss balanced physics-informed neural networks,'' {\em Neurocomputing}, vol.~496, pp.~11--34, 2022.

\bibitem{wang2022and}
S.~Wang, X.~Yu, and P.~Perdikaris, ``When and why pinns fail to train: A neural tangent kernel perspective,'' {\em Journal of Computational Physics}, vol.~449, p.~110768, 2022.

\bibitem{natarajan2014method}
V.~Natarajan and G.~Weiss, ``A method for proving the global stability of a synchronous generator connected to an infinite bus,'' in {\em 2014 IEEE 28th Convention of Electrical \& Electronics Engineers in Israel (IEEEI)}, pp.~1--5, IEEE, 2014.

\bibitem{wang2015power}
B.~Wang, Y.~Liu, and K.~Sun, ``Power system differential-algebraic equations,'' {\em arXiv preprint arXiv:1512.05185}, 2015.

\bibitem{kroutikova2007state}
N.~Kroutikova, C.~A. Hernandez-Aramburo, and T.~C. Green, ``State-space model of grid-connected inverters under current control mode,'' {\em IET Electric Power Applications}, vol.~1, no.~3, pp.~329--338, 2007.

\bibitem{du2020modeling}
W.~Du, F.~K. Tuffner, K.~P. Schneider, R.~H. Lasseter, J.~Xie, Z.~Chen, and B.~Bhattarai, ``Modeling of grid-forming and grid-following inverters for dynamic simulation of large-scale distribution systems,'' {\em IEEE Transactions on Power Delivery}, vol.~36, no.~4, pp.~2035--2045, 2020.

\bibitem{reuther2018interactive}
A.~Reuther, J.~Kepner, C.~Byun, S.~Samsi, W.~Arcand, D.~Bestor, B.~Bergeron, V.~Gadepally, M.~Houle, M.~Hubbell, M.~Jones, A.~Klein, L.~Milechin, J.~Mullen, A.~Prout, A.~Rosa, C.~Yee, and P.~Michaleas, ``Interactive supercomputing on 40,000 cores for machine learning and data analysis,'' in {\em 2018 IEEE High Performance extreme Computing Conference (HPEC)}, p.~1–6, IEEE, 2018.

\bibitem{majumder2024exploring}
S.~Majumder, L.~Dong, F.~Doudi, Y.~Cai, C.~Tian, D.~Kalathil, K.~Ding, A.~A. Thatte, N.~Li, and L.~Xie, ``Exploring the capabilities and limitations of large language models in the electric energy sector,'' {\em Joule}, vol.~8, no.~6, pp.~1544--1549, 2024.

\bibitem{chen2024gpt}
Y.~Chen and S.~Koohy, ``Gpt-pinn: Generative pre-trained physics-informed neural networks toward non-intrusive meta-learning of parametric pdes,'' {\em Finite Elements in Analysis and Design}, vol.~228, p.~104047, 2024.

\bibitem{raissi2017physics}
M.~Raissi, P.~Perdikaris, and G.~E. Karniadakis, ``Physics informed deep learning (part ii): Data-driven discovery of nonlinear partial differential equations,'' {\em arXiv preprint arXiv:1711.10566}, 2017.

\bibitem{misyris2021capturing}
G.~S. Misyris, J.~Stiasny, and S.~Chatzivasileiadis, ``Capturing power system dynamics by physics-informed neural networks and optimization,'' in {\em 2021 60th IEEE Conference on Decision and Control (CDC)}, pp.~4418--4423, IEEE, 2021.

\bibitem{stiasny2021physics}
J.~Stiasny, G.~S. Misyris, and S.~Chatzivasileiadis, ``Physics-informed neural networks for non-linear system identification for power system dynamics,'' in {\em 2021 IEEE Madrid PowerTech}, pp.~1--6, IEEE, 2021.

\bibitem{naglic2019pmu}
M.~Naglic, ``Pmu measurements of ieee 39-bus power system model,'' {\em IEEE Dataport}, 2019.

\bibitem{karniadakis2021physics}
G.~E. Karniadakis, I.~G. Kevrekidis, L.~Lu, P.~Perdikaris, S.~Wang, and L.~Yang, ``Physics-informed machine learning,'' {\em Nature Reviews Physics}, vol.~3, no.~6, pp.~422--440, 2021.

\bibitem{yang2021bv2}
L.~Yang, X.~Meng, and G.~E. Karniadakis, ``B-pinns: Bayesian physics-informed neural networks for forward and inverse pde problems with noisy data,'' {\em Journal of Computational Physics}, vol.~425, p.~109913, 2021.

\bibitem{stock2024bayesian}
S.~Stock, D.~Babazadeh, C.~Becker, and S.~Chatzivasileiadis, ``Bayesian physics-informed neural networks for system identification of inverter-dominated power systems,'' {\em arXiv preprint arXiv:2403.13602}, 2024.

\bibitem{radisavljevic2023modeling}
V.~Radisavljevic-Gajic, D.~Karagiannis, and Z.~Gajic, ``The modeling and control of (renewable) energy systems by partial differential equations—an overview,'' {\em Energies}, vol.~16, no.~24, p.~8042, 2023.

\end{thebibliography}
\newpage

\end{document}